\documentclass[conference]{IEEEtran}
\IEEEoverridecommandlockouts
\usepackage{cite}
\usepackage{amsmath,amssymb,amsfonts}
\usepackage{makecell}
\usepackage{algorithmic}
\usepackage{algorithm}
\usepackage{graphicx}
\usepackage{subfigure}
\usepackage{textcomp}
\usepackage{threeparttable} 
\usepackage{xcolor}
\usepackage{multirow}
\usepackage[colorlinks,
            linkcolor=black,
            urlcolor=blue,
            anchorcolor=black,
            citecolor=black
            ]{hyperref}
\usepackage[numbers,sort&compress]{natbib}
\def\BibTeX{{\rm B\kern-.05em{\sc i\kern-.025em b}\kern-.08em
    T\kern-.1667em\lower.7ex\hbox{E}\kern-.125emX}}
\begin{document}

\title{Consensus-Based Dynamic Task Allocation for Multi-Robot System Considering Payloads Consumption\\
 }


\DeclareRobustCommand*{\IEEEauthorrefmark}[1]{%
  \raisebox{0pt}[0pt][0pt]{\textsuperscript{\footnotesize #1}}%
}
\author{
    \IEEEauthorblockN{
        Xuekai Qiu\IEEEauthorrefmark{}, 
        Pengming Zhu\IEEEauthorrefmark{}, 
        Yiming Hu\IEEEauthorrefmark{}, 
        Zhiwen Zeng\IEEEauthorrefmark{}\textsuperscript{\textasteriskcentered},
        Huimin Lu\IEEEauthorrefmark{}
    }
    \IEEEauthorblockA{
        \IEEEauthorrefmark{}College of Intelligence Science and Technology, National University of Defense Technology, \\Changsha 410073, China\\
        Email: 
            zengzhiwen@nudt.edu.cn
    }
}


\maketitle

\begin{abstract}
This paper presents a consensus-based payload algorithm (CBPA) to deal with the condition of robots' capability decrease for multi-robot task allocation. During the execution of complex tasks, robots' capabilities could decrease with the consumption of payloads, which causes a problem that the robot coalition would not meet the tasks' requirements in real time. The proposed CBPA is an enhanced version of the consensus-based bundle algorithm (CBBA) and comprises two primary core phases: the payload bundle construction and consensus phases. In the payload bundle construction phase, CBPA introduces a payload assignment matrix to track the payloads carried by the robots and the demands of multi-robot tasks in real time. Then, robots share their respective payload assignment matrix in the consensus phase. These two phases are iterated to dynamically adjust the number of robots performing multi-robot tasks and the number of tasks each robot performs and obtain conflict-free results to ensure that the robot coalition meets the demand and completes all tasks as quickly as possible. Physical experiment shows that CBPA is appropriate in complex and dynamic scenarios where robots need to collaborate and task requirements are tightly coupled to the robots' payloads. Numerical experiments show that CBPA has higher total task gains than CBBA.
\end{abstract}

\begin{IEEEkeywords}
multi-robot system, task allocation, consensus-based, Payload Consumption
\end{IEEEkeywords}

\section{Introduction}
As artificial intelligence and robotics continue to advance, intelligent robots are playing an increasingly crucial role in various industries, including warehousing and logistics, search and rescue, and urban confrontation. However, expecting individual robots to handle all tasks may not be practical. This calls for considering multi-robot collaboration to tackle tasks effectively. In the theory of multi-robot collaboration, task allocation is of great research significance as it ensures that robots with appropriate capabilities are assigned to each task.

In the study of multi-robot task allocation (MRTA), there are various methods available, including optimization-based, learning-based and auction-based approaches. The optimization-based approachs\cite{gombolay2018fast,suslova2020multi,tehrani2022multi} model MRTA as an optimization problem. The optimization objectives include minimizing task time, task cost, and task revenue.
The learning-based approaches, such as deep Q-networks\cite{Lowe2017Multi,gong2022weapon} and graph neural networks\cite{tolstaya2021multi,paul2023efficient}, are used to enhance task allocation effectiveness. 
The auction-based methods\cite{choi2009consensus,hunt2014consensus,li2020dynamic,ye2021decentralized} are an approximation of market bidding mechanism, where robots obtain the execution rights of tasks through bidding. 
The consensus-based bundle algorithm (CBBA)\cite{choi2009consensus}, an auction-based method, is a distributed two-stage iterative consensus algorithm with good scalability and dynamic characteristics. However, this algorithm addresses the single-robot task\cite{gerkey2004formal} allocation problem and does not extend to multi-robot tasks requiring multiple robots' combined effort.

In recent years, many researchers have worked on improving CBBA to solve the task allocation problem under complex coupling constraints and requiring collaboration between multiple robots.
The consensus-based grouping algorithm (CBGA)\cite{hunt2014consensus} expands the winner list and bids list of each robot in CBBA into a two-dimensional vector to solve the allocation problem of multi-robot tasks. 
The consensus-based coalition algorithm (CBCA)\cite{li2020dynamic} investigates the task allocation problem under task payload resource constraints, sub-task coupling relationship constraints, and execution window constraints. 
The extended CBBA with task coupling constraints (CBBA-TCC)\cite{ye2021decentralized} algorithm
clarifies each robot's tasks to cope with task coupling constraints in heterogeneous multi-robot system.
However, both CBCA and CBBA-TCC
can't cope with the problem of robot payload resource consumption.
A hierarchy-based extension of the consistent bundle algorithm \cite{liu2022hierarchy} investigates the MRTA problem under complex constraints, but the effect of the robots' arrival time on the start time of multi-robot tasks needed to be clarified.
The consensus-based timetable algorithm (CBTA)\cite{wang2022consensus} 
argues that the arrival time of the last robot determines the start time of a multi-robot task.
However, the number of tasks a robot can perform is not dynamically adjusted to its capabilities in CBTA.
It is worth noting that during the execution of a multi-robot task, the robots are usually heterogeneous. Simultaneously, the robots' capabilities to execute tasks may be decreased.
For example, in military confrontation, the strike payload carried by robots will be consumed during the execution of attack tasks, and the robot's strike capability will also be weakened accordingly.

This paper presents a consensus-based payload algorithm (CBPA) to address the MRTA problem when robots' capabilities decrease due to payload consumption during task execution. 
The CBPA framework consists of two main phases: the payload bundle construction and consensus phases. 
In the payload bundle construction phase, CBPA introduces a payload assignment matrix to keep track of the payloads carried by the robots and the demands of multi-robot tasks in real time. It dynamically adjusts the number of robots performing multi-robot tasks and the number of tasks each robot performs, ensuring that all tasks are completed as quickly as possible by the robot coalition that meets the demand. 
In the consensus phase, the paper establishes consensus rules based on different task requirements and explains the convergence, ensuring a conflict-free task allocation result. 
In a physical multi-robot system, it is demonstrated that CBPA has excellent resource utilization and can fully utilize the task execution capability. Experiments show that CBPA has higher total task gains than CBBA.


\section{Problem Description}
We focus on allocating multi-robot tasks with known positions and demands to multi-robot system carrying different task payloads.
Robots performing the same task form a robot coalition, and demonstrate their adaptability by forming a new coalition for the next task until all tasks are completed or they are not able to participate due to payload decrease. Each coalition's payload mix must meet the tasking requirements.

To describe the above task allocation problem in a mathematical expression, we denote the robots set by \(I=\left\{ 1,2,...,N_r \right\} \), where $N_r$ is the number of robots.
The attributes of robot $i\in I$ are described as $\left< p_i,v_i,\tilde{l}^a_i,\tilde{l}^b_i \right>$, where $p_i$ is the position of robot $i$, $v_i$ is the velocity, and $\tilde{l}^a_i$ and $\tilde{l}^b_i$ are two different payloads, payload A and B, carried by robot $i$.
In the urban confrontation payload A and B can be represented as reconnaissance and strike payloads for ground robots.
The set of tasks is denoted by $J=\left\{ 1,2,...,N_t \right\}$, where $N_t$ is the number of tasks. The attributes for task $j$ are described as $\left< p'_j,t'_j,\delta _j,\tilde{r}^a_j,\tilde{r}^b_j \right>$, where $p'_j$ is the position of task $j$, $t'_j$ is the duration, $\delta _j$ is the execution time, and $\tilde{r}^a_j$ and $\tilde{r}^b_j$ are the payload requirements of task $j$ for different functional types.

For robot $i$, the $N_t\times N_r$ matrix $L^a_i$ is defined to describe the assignment of payload A during the task allocation process.
The element $l^a_{ijk}\in L^a_i$ denotes the value of payload A assigned to robot $k$ to execute task $j$ in robot $i$'s perspective. If robot $k$ does not execute task $j$, then $l^a_{ijk}=0$. Similarly, the element $l^b_{ijk}\in L^b_i$ denotes the assignment of payload B.
In addition, to keep track of the remaining payloads of robots after completing a series of tasks, lists $\boldsymbol{r}^a_i$ and $\boldsymbol{r}^b_i$ of length $N_r$ denote the amount of payload A and B remaining for robot $i$. The element $r^a_{ik}$ and $r^b_{ik}$ denote the remaining payloads of robot $k$ in robot $i$'s perspective. 
It is worth emphasizing that the payload remainder can be calculated from the robots' payload assignment.
If payload A is consumable, such as the strike payload in military confrontation, then 
\begin{equation}
    r^a_{ik}=\tilde{l}^a_k-\sum_{m=1}^{N_t}{l^a_{imk}}.
\end{equation}
If payload B is non-consumable during the execution of the task, such as the reconnaissance payload in military confrontation, the payload remainder is equal to the initial robot payload, i.e.
\begin{equation}
    r^b_{ik}=\tilde{l}^b_k.
\end{equation}

To determine whether the requirements of task $j$ are met, task requirements lists $M^a_i$ and $M^b_i$ of length $N_t$ are defined.
The elements $m^a_{ij}$ and $m^b_{ij}$ denote the dynamic requirements of task $j$ for payload A and B in robot $i$'s perspective, which are calculated as 
\begin{equation}
    m_{ij}=\tilde{r}_j-\sum_{k=1}^{N_r}{l_{ijk}},
    \label{Eq:3}
\end{equation}
when $m_{ij}=0$, it means that task $j$ satisfies the task requirements.

While the task allocation is completed, robot $i$ forms a task execution path $\boldsymbol{p}_i=\left\{ p_{i1},p_{i2},...,p_{i|p_i|} \right\} $, where $p_{ik}$ denotes the k-th task executed by robot $i$ and $|\boldsymbol{p}_i|$ is the number of tasks assigned to robot $i$.
Task $j$ starts only when the robot coalition executing task $j$ meets the requirements and reaches the task position. 
Therefore, tasks' start times are determined by the last arriving robot, and the start time of task $j$ is denoted as 
\begin{equation}
\tau _j=\left\{ \begin{array}{ll}
	\max_{}t_{ijk}\cdot x_{ijk},&\text{if\,\,}\sum_{k=1}^{N_r}{l_{ijk}\ge\tilde{r}_{j}}\\
	\infty,&\text{otherwise}\end{array} \right.,
 \label{Eq:4}
\end{equation}
where $t_{ijk}$ denotes the time that robot $i$ thinks robot $k$ arrives at task $j$.
If the robot coalition executing the task does not meet the requirement, the start time is $\infty$.

Since the timeliness of task execution is critical in highly dynamic scenarios, multi-robot system need to accomplish tasks quickly and efficiently to address new tasks.
Therefore, the optimization objective for task allocation is to achieve the fastest average task start time. It is described as the following optimization problem:
\begin{equation}
\setlength{\abovedisplayskip}{5pt}
\setlength{\belowdisplayskip}{5pt}
    \min \left\{ \frac{1}{N_r}\left( \sum_{j=1}^{N_t}{\tau _j} \right) \right\},
\end{equation}
s.t.
\begin{equation}
    \sum_k^{N_r}{l_{ijk}\ge \tilde{r}_j},\ \forall j\in J,
\label{Eq:6}
\end{equation}
\begin{equation}
    \sum_j^{N_t}{l_{ijk}\le \tilde{l}_k,\ \forall k\in I},
\label{Eq:7}
\end{equation}
\begin{equation}
    L_i=L_k,\ \forall i,k\in I,
\label{Eq:8}
\end{equation}
where the various requirements of the task are satisfied in \eqref{Eq:6}, the robots are capable of executing the assigned task in \eqref{Eq:7}, and the payload assignment scheme is consistent among the robots in \eqref{Eq:8} to avoid conflicts.

\section{Consensus-Based Payloads Allocation Algorithm}
CBPA is an enhanced version of CBBA designed to tackle issues in complex and dynamic scenarios where robots need to collaborate and task requirements are tightly coupled to the robots' payloads.
The framework of CBPA comprises two primary core phases: the payload bundle construction and consensus phases. These two phases run distributed on robots and alternate to form an iterative process to ensure that tasks are allocated to each robot in a productive, precise, and conflict-free manner.

The following are the key elements of CBPA:
\begin{enumerate}
    \item[1)]Winning Robots Matrix $\textbf{X}_i$
\end{enumerate}

$\textbf{X}_i=\{ x_{ijk}\in \left\{ 1,0 \right\} ;\ \forall \left( j,k \right) \in J\times I\}$ is a matrix sized $N_t\times N_r$, where $x_{ijk}=1$ denotes that robot $i$ thinks that robot $j$ executes the task $j$, otherwise $x_{ijk}=0$.

\begin{enumerate}
    \item [2)]Robot Time Matrix $\textbf{T}_i$
\end{enumerate}

$\textbf{T}_i=\{ t_{ijk}\in \left[ -1,\infty \right) ;\ \forall \left( j,k \right) \in J\times I \}$ is a matrix sized $N_t\times N_r$, where $t_{ijk}>0$ denotes the time when robot $i$ thinks that robot $k$ gets to task $j$. When $t_{ijk}=-1$, it means that robot $k$ does not execute task $j$, and $x_{ijk}=0$.

\begin{enumerate}
    \item [3)] Payload Assignment Matrix $\textbf{L}_i$
\end{enumerate}

The payload assignment of robot i to execute task j is based on the remaining payloads of robots and the requirements of the task. We assume an equal distribution of payloads among the robot coalition. Each member of the robot task coalition consumes the same payload value per unit of time, which is the simplest and most common model of payload assignment. For example, similar combat robots on the battlefield consume almost the same amount of ammunition per unit of time. The payload assignment for task j in a coalition can be calculated using Algorithm 1. Similar algorithms can be developed for other complex payload allocation models.

\begin{enumerate}
    \item [4)] Task Bundle $\boldsymbol{b}_i$
\end{enumerate}

$\boldsymbol{b}_i=\left\{ b_{i1},b_{i2},...,b_{i|b_i|} \right\}$ is a list of non-fixed length, $\boldsymbol{b}_i\subseteq J$, and denotes the set of tasks allocated to robot $i$. The task execution path $\boldsymbol{p}_i$ is obtained by sorting the elements in $\boldsymbol{b}_i$ according to robot $i$'s arrival time order at the task.  

\begin{enumerate}
    \item [5)] Timestamp List $\boldsymbol{s}_i$
\end{enumerate}

The timestamp list $\boldsymbol{s}_i$ denotes the updated time of the current task allocation information, which is updated in each iteration and serves as an important basis for judgment in the consensus phase. $s_{ij}$ denotes the time when robot $i$ updates task $j$.

\begin{algorithm}[!h]
    \caption{Payload Average Allocation}
    \label{A1}
    \renewcommand{\algorithmicrequire}{\textbf{Input:}}
    \renewcommand{\algorithmicensure}{\textbf{Output:}}
    \begin{algorithmic}[1]
        \REQUIRE Winners list of task $j$: $\textbf{x}_{ij}$, remaining payload of robots: $r_i$, requirements of task $j$: $\tilde{r}_j$  
        \ENSURE Payload assignment matrix $\boldsymbol{l}_{ij}$    
        
        \STATE  $r'_i=\sum{\left( \boldsymbol{r}_i\cdot \boldsymbol{x}_{ij} \right)}$, $\boldsymbol{l}_{ij}$
        
        \IF {$r'_i\le \tilde{r}_j$}
            \STATE $\boldsymbol{l}_{ij}=\boldsymbol{r}_i$
            \ELSE
            \WHILE {$\tilde{r}_j$}
                \STATE $\bar{l}=\tilde{r}_j/\text{len}(\{ x_{ijk}|x_{ijk}=1,k\in I \}) $
                \IF {$l_{ijk}<r_{ii}\le \bar{l}$}
                    \STATE $l_{ijk}=r_{ii},\ \tilde{r}_j=\tilde{r}_j-r_{ii},\ x_{ijk}=0$
                \ELSE
                    \STATE $l_{ijk}=l_{ijk}+\bar{l},\ \tilde{r}_j=\tilde{r}_j-\bar{l}$
                \ENDIF
            \ENDWHILE
        \ENDIF
        
        
        \RETURN $\boldsymbol{l}_{ij}$ 
    \end{algorithmic}
\end{algorithm}

\subsection{Payload Bundle Construction Phase}\label{AA}
In the payload bundle construction phase, the robot select tasks based on the requirements of the tasks and its own capabilities. This process is synchronized for each robot. 
During the bundle construction of robot $i$, robot $i$ first updates its winners matrix based on the latest robots time matrix. If $t_{ijk}<N$, then $x_{ijk}=1$, where $N$ is a sufficiently large number to ensure that all tasks are executed. Tasks that lose their bids will be removed.

Robot $i$ then bids on tasks that match its capability and selects the task with the lowest incremental marginal cost $C_{ij}$ to add to its bundle. For the marginal cost of task $j$, robot $i$ inserts the task into $\boldsymbol{p}_i$ and chooses the insertion position with the smallest increment, i.e.
\begin{equation}
    C_{ij}=\left\{ \begin{array}{l}
	c,\,\,\ \ \ \ \ \ \ \ \ \  r_{ii}>0\,\,\text{and\,\,}\tilde{r}_j>0\\
	\min_{n\le \left| \boldsymbol{p}_i \right|+1}\left\{ S_i\left( \boldsymbol{p}_i\oplus _n\left\{ j \right\} \right) -S_i\left( \boldsymbol{p}_i \right) \right\}\\
\end{array} \right.,
\label{Eq:9}
\end{equation}
where $S_i(\boldsymbol{p}_i)$ denotes the marginal cost of robot $i$ along the task path $\boldsymbol{p}_i$; $\boldsymbol{p}_i\oplus _n\{ j\}$ denotes the insertion of task $j$ into the n-th position in $\boldsymbol{p}_i$; and $c$ is a sufficiently large constant that denotes the cost when the robot not meets the task requirements.

The marginal cost $S_i$ of robot $i$ is defined as
\begin{equation}
    S_i(\boldsymbol{p}_i) =\sum_{j=1}^{N_t}{\left[ \left( \alpha \cdot m^a_{ij}+\beta \cdot m^b_{ij}+t_{iji} \right) x_{ijk} \right]},
\end{equation}
where $m^a_{ij}$ and $m^b_{ij}$ represent the remaining payloads for task $j$ in robot $i$'s perspective and can be calculated from (\ref{mij}). When the requirement for payload A is met, $m^a_{ij}=0$; the situation is similar for payload B. $\alpha$ and $\beta$ are sufficiently large numbers that robot $i$ will prioritize tasks that have not yet met the requirement and denote the weights of the different payload requirements to differentiate between the priorities of different tasks.
\begin{equation}
    m_{ij}=\tilde{r}_{j}^{}-\sum_{k=1}^{N_r}{l_{ijk}}
    \label{mij}
\end{equation} 

The time for robot $i$ to reach task $j$ is determined by the completion time of the previous task and the distance of the current task. Assuming task $j$ is the $k$-th task in $\boldsymbol{p}_i$, $t_{iji}$ is calculated as
\begin{equation}
    t_{i,p_{ik},i}=\left\{ \begin{array}{ll}
	\frac{d_{i,p_{ik}}}{v_i},& k=0\\
        \tau _{i,p_{i,k-1}}+\delta _{p_{i,k-1}}+\frac{d_{p_{i,k-1},p_{ik}}}{v_i},& k\ne 0\\
\end{array}\right.
\end{equation}
where $d_{i,p_{ik}}$ denotes the distance between robot $i$ and the $k$-th task in $\boldsymbol{p}_i$, $v_i$ is the velocity of robot $i$ and $\delta$ denotes the task duration.

Finally, the payload bundle construction of robot $i$ continues until there are no unassigned tasks or robot $i$ is under-capable, as shown in Algorithm \ref{A2}.
\begin{algorithm}[!h]
    \caption{Payload Bundle Construction of Robot $i$}
    \label{A2}
    \renewcommand{\algorithmicrequire}{\textbf{Input:}}
    \renewcommand{\algorithmicensure}{\textbf{Output:}}
    \begin{algorithmic}[1]
        \REQUIRE Robot time matrix $\textbf{T}_i(t-1)$, payload assignment matrix $\textbf{L}^a_i(t-1)$, $\textbf{L}^b_i(t-1)$  
        \ENSURE $\textbf{T}_i(t),\ \textbf{L}^a_i(t),\ \textbf{L}^b_i(t) $    
        
        \STATE Update $\boldsymbol{X}_i,\ \boldsymbol{r}_i,\ \boldsymbol{M}_i,\boldsymbol{\tau }_i,\ \boldsymbol{b}_i,\ \boldsymbol{p}_i$
        \WHILE{$r^a_{ii}>0$ or $r^b_{ii}>0$}
            \STATE Compute $C_{ij},\ \forall j\in J$
            \STATE $g=\min _{j\le N_t}C_{ij}$
            \IF{$0<g<c$}
                \STATE $q=\text{arg}\min _{j\le N_t}C_{ij}$
                \STATE $h=\text{arg}\min _{n\le \left| \boldsymbol{p}_i \right|+1}\left\{ S_i\left( \boldsymbol{p}_i\oplus _n\left\{ q \right\} \right) \right. \left. -S_i\left( \boldsymbol{p}_{\boldsymbol{i}} \right) \right\}$
                \IF{$m^a_{iq}>0$ or $m^b_{iq}>0$}
                    \STATE $x_{iqi}=1,\ t_{iqi}=t_{i,p_{ih},i}$
                    \STATE Compute and update $\textbf{L}_i$ according to algorithm \ref{A1}
                \ENDIF
                \IF{$m^a_{iq}=0$ or $m^b_{iq}=0$}
                    \STATE $i'=\text{arg}\max _{x_{iqk}=1}t_{iqk}\cdot x_{iqk}$
                    \STATE $x_{iqi}=1,\ x_{iqi'}=0$
                    Compute $\textbf{L}_i$
                    \IF{$m^a_{iq}=0$ or $m^b_{iq}=0$}
                        \STATE $t_{iqi}=t_{i,p_{ih},i}$, and update $\textbf{L}_i$
                    \ENDIF
                \ENDIF
                \STATE Update $M_i,\ \boldsymbol{\tau }_i,\ \boldsymbol{b}_i,\ \boldsymbol{p}_i$
            \ENDIF
        \ENDWHILE
    \end{algorithmic}
\end{algorithm}

Algorithm \ref{A2} will run as each robot builds its payload bundle. The robots will choose their respective optimal task $q$ (line 6), determine whether task $q$ has been met, and prioritize the allocation of tasks that do not meet the task requirement (line 8). 
If met, the robots will replace the slowest robot $i'$ to reduce the task start execution time (line 9).

\subsection{Consensus Phase}
multi-robot system need to share robot and task information through the current network setup to update the information that robots need in the next round of bidding, such as the remaining payload of robots and the matching of task requirements. Additionally, different robots may have selected the same task in one round of bidding, causing conflicts in task allocation, so the consensus phase focuses on information sharing and conflict resolution.

The consensus phase runs independently in each robot, and robots share the task time matrix, payload assignment matrix, and timestamp information. Robot $i$ sends $\textbf{T}_i$, $\textbf{L}_i$ and $\boldsymbol{s}_i$ to robot $k$ with which it has established a communication connection, i.e. $g_{ik}$, where $G$ denotes the network topology matrix of a multi-robot system.
Robot $i$ also receives $\textbf{T}_b$, $\textbf{L}_b$ and $\boldsymbol{s}_b$ sent by robot $k$ and calculates the start time $\tau _{kj}$ and demand fulfillment for each task $m_{kj}$ in robot $k$'s perspective according to \eqref{Eq:3} and \eqref{Eq:4}.
Then robot $i$ compares the task allocation information of robot $k$ and updates $\textbf{T}_i$ and $\textbf{L}_i$, i.e. $\textbf{T}_{ij}=\textbf{T}_{kj}$, $\textbf{L}_{ij}=\textbf{L}_{kj}$ and $\boldsymbol{s}_i=\boldsymbol{s}_k$, based on the rules in TABLE \ref{T1}.
$\textbf{X}_i$, $\boldsymbol{r}_i$, $\textbf{M}_i$, $\boldsymbol{\tau}_i$, $\boldsymbol{b}_i$ and $\boldsymbol{p}_i$ will be updated during the next round of bundle construction.
\begin{table}[htbp]
\renewcommand\arraystretch{1.1}
\caption{Update Rules for Task $j$ in Robot $i$\\(received from robot k)}
\begin{center}
\begin{tabular}{c|c|c}
\Xhline{0.8pt}
$\boldsymbol{t}_{iji}$&$\boldsymbol{t}_{kjk}$&\textbf{Condiction of Update}\\
\Xhline{0.8pt}
-1&-1&$\boldsymbol{s}_{ij}<\boldsymbol{s}_{kj}$\\
\hline
\multirow{2}{*}{-1}&\multirow{2}{*}{$[ 0, \ \infty)$}&$\boldsymbol{\tau}_{ij}>\boldsymbol{\tau}_{kj}$\\
\cline{3-3}
 & & $\boldsymbol{\tau}_{ij}=\boldsymbol{\tau} _{kj}\,\,\text{and\,\,}\boldsymbol{m}_{ij}\ge \boldsymbol{m}_{kj}$\\
\hline
\multirow{2}{*}{$[ 0, \ \infty)$}&\multirow{2}{*}{-1}&$\boldsymbol{\tau}_{ij}>\boldsymbol{\tau}_{kj}\,\,\text{and\,\,}\boldsymbol{s}_{ij}<\boldsymbol{s}_{kj}$\\
\cline{3-3}
 & & $\boldsymbol{\tau}_{ij}=\boldsymbol{\tau} _{kj},\ \boldsymbol{m}_{ij}\ge \boldsymbol{m}_{kj}\,\, \text{and\,\,}\boldsymbol{s}_{ij}<\boldsymbol{s}_{kj}$\\
\hline
\multirow{2}{*}{$[ 0, \ \infty)$}&\multirow{2}{*}{$[ 0, \ \infty)$}&$\boldsymbol{\tau}_{ij}>\boldsymbol{\tau}_{kj}$\\
\cline{3-3}
 & & $\boldsymbol{\tau}_{ij}=\boldsymbol{\tau} _{kj}\,\,\text{and\,\,}\boldsymbol{m}_{ij}\ge \boldsymbol{m}_{kj}$\\
\hline
\end{tabular}
\label{T1}
\end{center}
\end{table}

\subsection{Convergence}
Based on the task update rules in TABLE \ref{T1}, in each round of bidding, the robot that meets the requirements of task $j$ will win first and get the right to execute task $j$. When task requirements are met, i.e. $m_{ij}=0$ and $t_{ij}\ll \infty$, then the n-th round of iteration update with
\begin{equation}
    \left\{ \begin{array}{l}
	m_{ij}^{\left( n+1 \right)}\le m_{ij}^{\left( n \right)}\\
	t_{ij}^{\left( n+1 \right)}\le t_{ij}^{\left( n \right)}\\
\end{array} \right. .
\nonumber
\end{equation}

If no robot meets the task requirements, i.e. each robot starts the task $j$ simultaneously and tends to infinity, the robot with the smallest task requirements will win. Then the requirements of the updated task will be smaller, i.e. 
\begin{equation}
    m_{ij}^{\left( n+1 \right)}\le m_{ij}^{\left( n \right)}.
    \nonumber
\end{equation}

If all robots think that task $j$'s requirements have been satisfied, the robot with the smallest start time wins.
Then the updated start time of the task will be smaller, i.e.
\begin{equation}
    t_{ij}^{\left( n+1 \right)}\le t_{ij}^{\left( n \right)}.
    \nonumber
\end{equation}

In line 9 of the bundle construction algorithm, task $j$ will only be added to the robot's task bundle if the robot can reduce the task start time. Combining the above three scenarios, after each round of consensus, the value added of the marginal cost calculated by the robot according to \eqref{Eq:9} has a decreasing trend, i.e. 
\begin{equation}
    C_{ij}^{\left( n+1 \right)}\le C_{ij}^{\left( n \right)}.
\end{equation}
When $x_{iji}=0$, $C_{ij}^{\left( n+1 \right)}=0$. When the task is not updated, $C_{ij}^{\left( n+1 \right)}=C_{ij}^{\left( n \right)}$.
This demonstrates that it is a condition of diminishing marginal gain (DMG).

Furthermore, the multi-robot communication network considered in this paper is static, and the connections between the robots are undirected. Thus, the two phases of CBPA converge in finite time as long as there is at least one feasible communication path in each of the two robots\cite{choi2009consensus}.

\begin{table*}[!htbp]
\renewcommand\arraystretch{1}
\caption{Initial Positions and Payload Information in Case 1}
\begin{center}
\begin{threeparttable} 
\begin{tabular}{ccccccccc}
\Xhline{0.8pt}
&\textbf{R1}\tnote{1}&\textbf{R2}&\textbf{R3}&\textbf{R4}&\textbf{R5}&\textbf{T1}\tnote{1}&\textbf{T2}&\textbf{T3}\\
\hline 
$\textbf{P}\tnote{2}$ & (150,130) & (150,330) & (150,700) & (150,970) & (150,1270) & (850,350) & (1450,550) & (2100,430) \\
\hline
$\boldsymbol{L}_a/\boldsymbol{r}_a$\tnote{3} & 0 & 0 & 30 & 30 & 25 & 8 & 6 & 8\\
\hline
$\boldsymbol{L}_b/\boldsymbol{r}_b$\tnote{4} & 3 & 3 & 3 & 0 & 0 & 1 & 2 & 3\\
\Xhline{0.8pt}
&\textbf{T4}&\textbf{T5}&\textbf{T6}&\textbf{T7}&\textbf{T8}&\textbf{T9}&\textbf{T10}&\\
\hline
$\textbf{P}$ & (740,820) & (1120,740) & (1710,970) & (2360,680) & (730,1260) & (1440,1020) & (1950,1280) & \\
\hline
$\boldsymbol{r}_a$ & 7 & 8 & 6 & 9 & 10 & 8 & 9 & \\
\hline
$\boldsymbol{r}_b$ & 3 & 3 & 2 & 2 & 3 & 2 & 2 & \\
\Xhline{0.8pt}
\end{tabular}
\begin{tablenotes}
        \footnotesize 
        \item[1] Identification of robots and tasks.\item[2] Position of robots and tasks in meters.
        \item[3] Strike payloads of robots and Strike requirements of tasks.
        \item[4] Reconnaissance  payloads of robots and reconnaissance requirements of tasks.
\end{tablenotes}    
\end{threeparttable} 
\label{T2}
\end{center}
\end{table*}
\section{Experiments and Results }
We utilize simulation combined with physical robots to conduct experiments. 
The effectiveness and feasibility of the proposed CBPA are confirmed by physical experiments on dynamic task allocation.
The task revenue of the proposed algorithm are evaluated by comparing it with CBBA in extensive simulations.
\subsection{Case 1: Algorithm Performance Verification}
To verify the effectiveness and feasibility of the algorithm, we design Case 1, which involves a mobile robot task allocation problem in an urban adversarial environment. In Case 1, ten multi-robot tasks are executed by five robots equipped with reconnaissance and strike payloads within a task area measuring 2.4 km × 1.5 km. The reconnaissance payloads are non-expendable, while the strike payloads are expendable. Numerically simulated ammunition quantities represent the value of the strike payloads, and the ammunition consumption within ten is used to determine the strike payload requirements of the tasks. Each task requires the participation of at least one robot equipped with a reconnaissance payload.
TABLE \ref{T2} lists the initial positions and payload information for robots and tasks.

\begin{figure}[!ht]
    \centering
    \subfigure[CBPA]{\includegraphics[width=.24\textwidth]{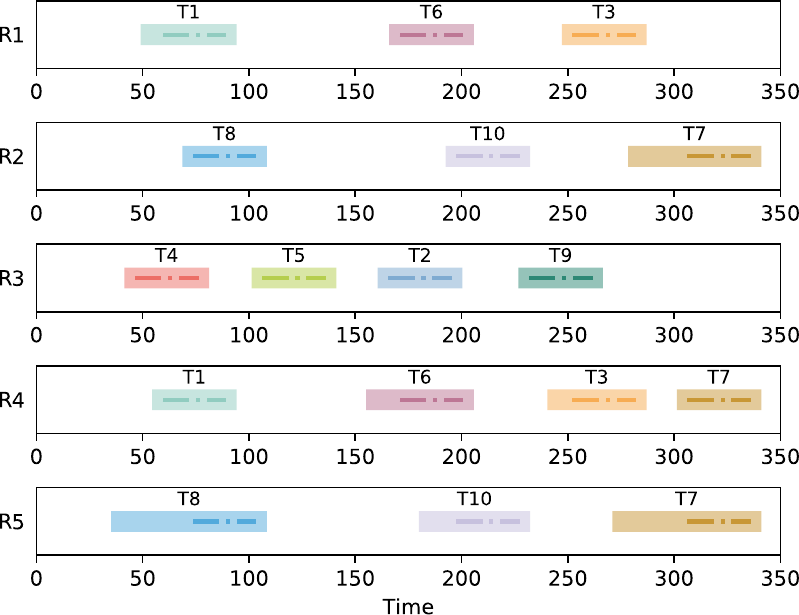}}
    \subfigure[Auction-Based Algorithm]{\includegraphics[width=.24\textwidth]{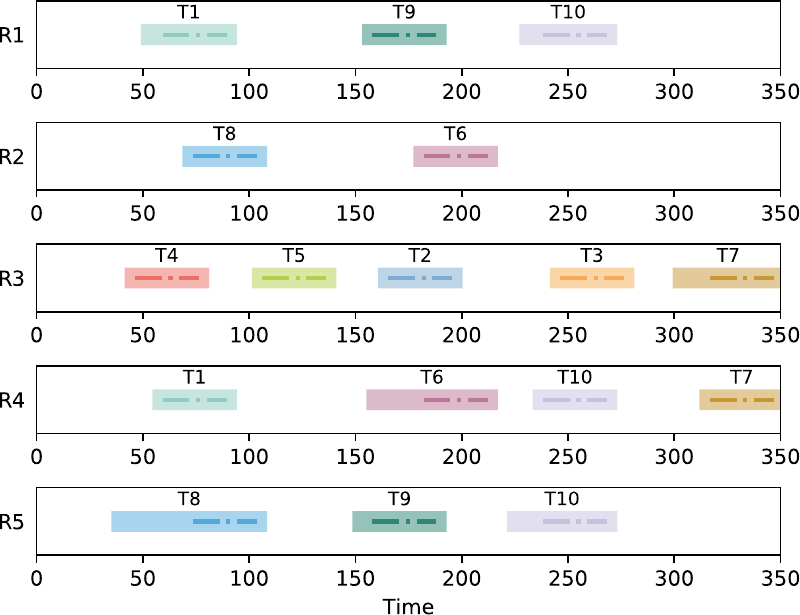}}
 \caption{Task Schedules}
 \vspace{-0.5cm}
 \label{F1}
\end{figure}

With Case 1, we thoroughly compare the calculation time and the average start time of the task between CBPA and the auction-based algorithm. The results are demonstrated with physical objects to verify the effectiveness and feasibility of CBPA, providing a reliable basis for further research and application.
The auction-based algorithm involves robots bidding on only one task per round and assigning the task to the best robot.
The results of the task allocation are depicted in Fig.~\ref{F1}.
The calculation time and average start time of tasks for CBPA and auction-based algorithm are shown in TABLE \ref{T3}.
CBPA has fewer iterations and shorter calculation time, i.e., it takes less time for task allocation and has better rapidity. This is because in CBPA the robot bids on multiple tasks in each round of bidding based on the current capabilities of robots.
\begin{table}[!ht]
\renewcommand\arraystretch{1}
\caption{Calculation Time and Average Start Time of Tasks}
\begin{center}
\begin{tabular}{cccc}
\Xhline{0.8pt}
              & \textbf{Iterations} & \textbf{Calculation time} & \textbf{Average time} \\ \Xhline{0.8pt}
\textbf{CBPA} & 12                  & 0.042993                  & 160.913218            \\ \hline
\textbf{AOA}  & 16                  & 0.049136                  & 159.239914            \\ \hline
\end{tabular}
\label{T3}
\end{center}
\vspace{-0.5cm}
\end{table}

\begin{figure}[h]
    \centering
    \includegraphics[width=2.6in]{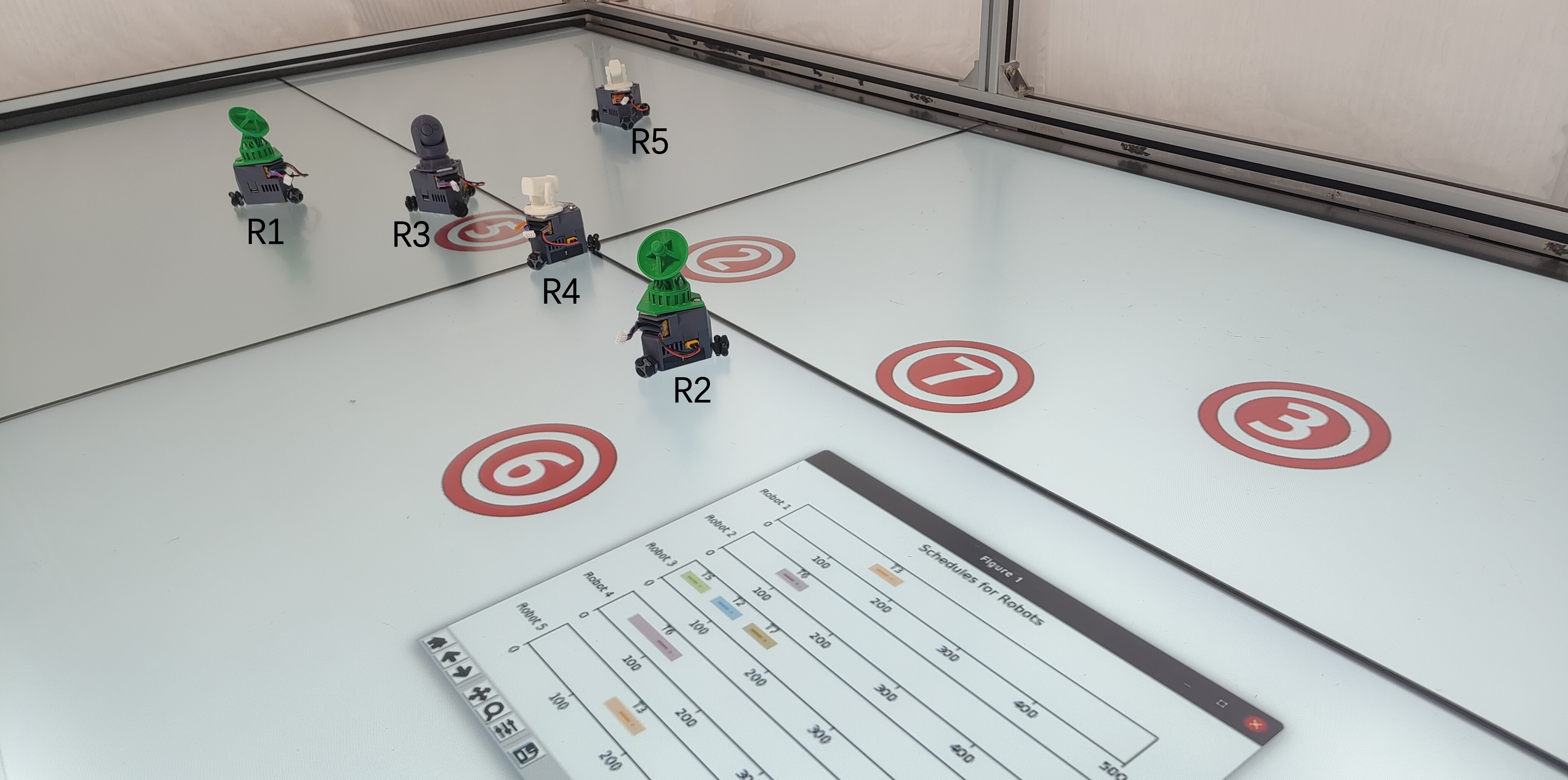}
    \caption{The mobile multi-robot physical platform of MRTA. Green robots carry reconnaissance payload, white robots carry strike payload, and gray robot carries reconnaissance and strike payloads}
    \label{F6}
\end{figure}

\begin{figure}[!ht]
\centering
\includegraphics[width=2.7in]{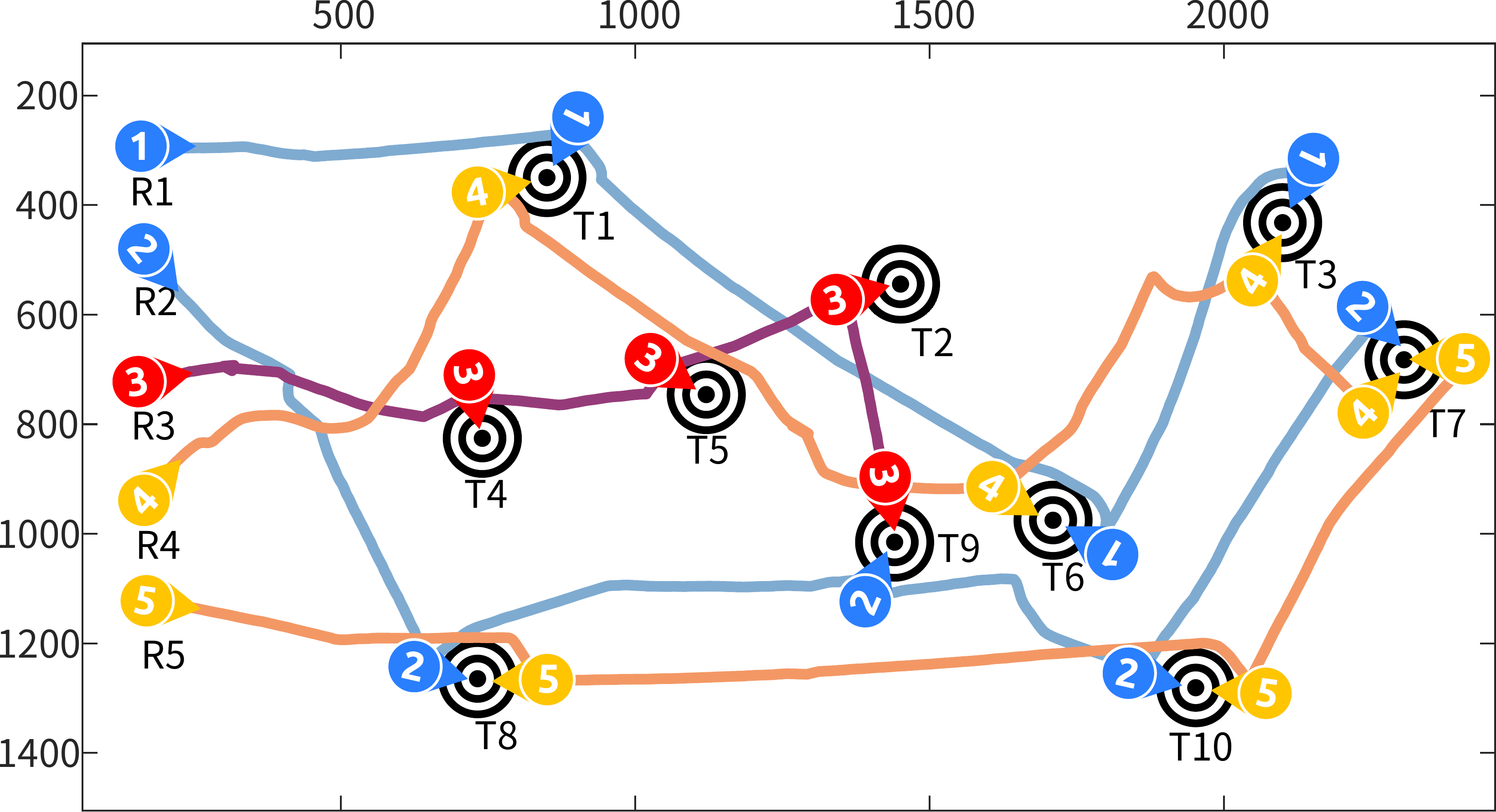}
\caption{Task Execution Path of CBPA}
\label{F3}
\end{figure}

The allocation results were verified on a mobile multi-robot physical platform (as shown in Fig.\ref{F6}). 
The tasks execution paths of robots are shown in Fig.\ref{F3}. Black circles denote successfully executed tasks. The result shows that CBPA has good effectiveness and feasibility in the application of physical systems.

\begin{figure}[!ht]
\centering
\includegraphics[width=2.7in]{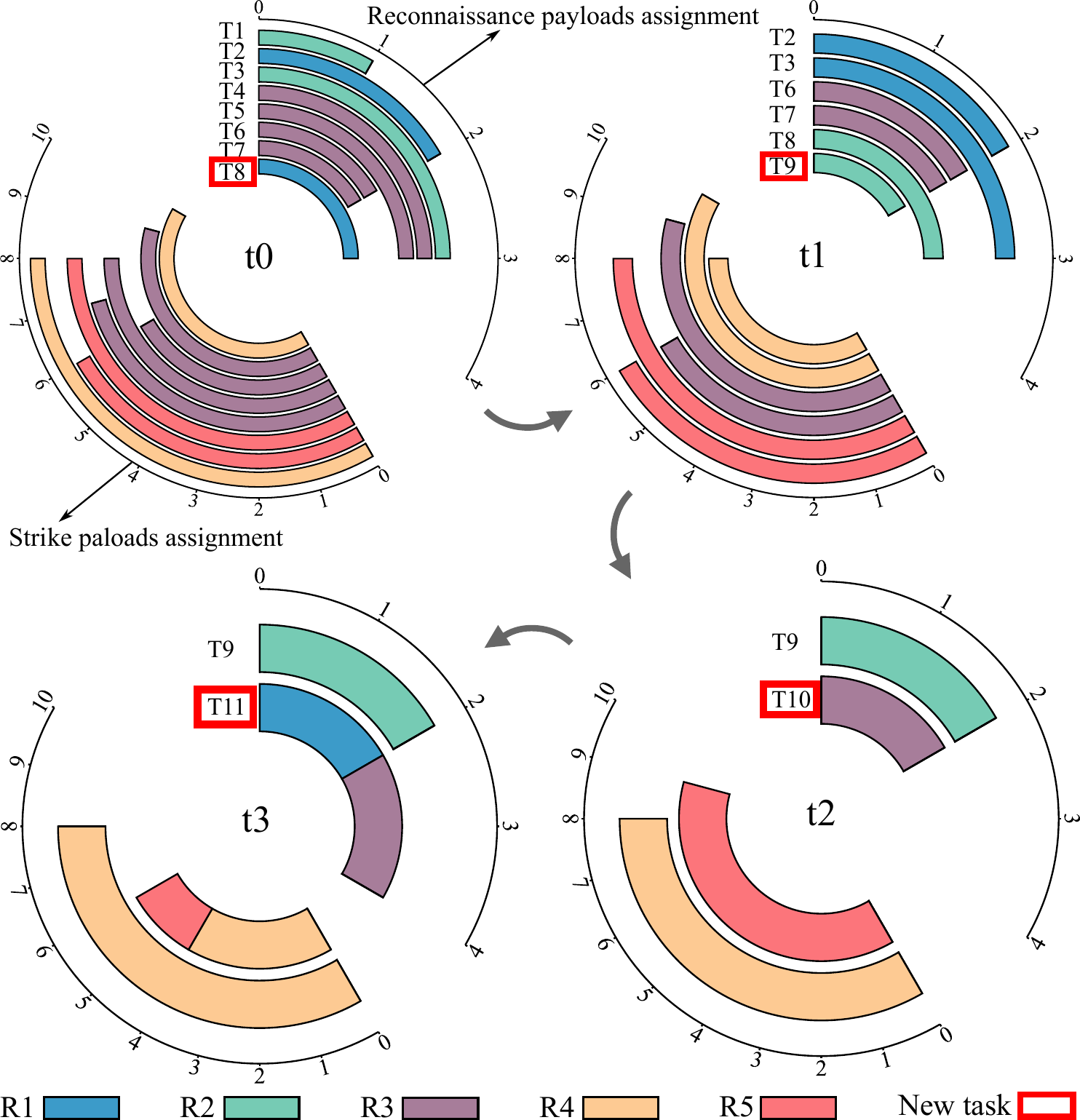}
\caption{Task Execution Path of Auction-based Algorithm}
\label{F4}
\end{figure}

To further verify the feasibility of CBPA in dynamic scenarios, tasks 1-8 in Case 1 are set as known tasks, and new tasks 9-11 are randomly introduced during the execution of the tasks, which is showed at \href{https://www.bilibili.com/video/BV1Q6vTeHEgg/?spm_id_from=333.999.0.0}{https://www.bilibili.com/video/BV1Q6vTeHEgg}. The results can be seen in Fig.\ref{F4}, where new tasks 9-11 occur at times t1-t3. When new tasks are identified, CBPA can dynamically adjust the executing robots according to current robots' capabilities.




\subsection{Case 2: Result Gains Evaluation}
In order to evaluate the results of task allocation, we compare the per-task gain of the proposed CBPA with that of the CBBA, which is computed as
\begin{equation}
    \omega =\omega _0\cdot \frac{\tilde{r}_j-m_j}{\tilde{r}_j}\cdot e^{-\lambda \cdot \tau _j},
    \label{value}
\end{equation}
where $\omega_0$ denotes the static gain of the task and $\lambda$ is the time discount factor.

\begin{figure}[!ht]
\centering
\includegraphics[width=2.6in]{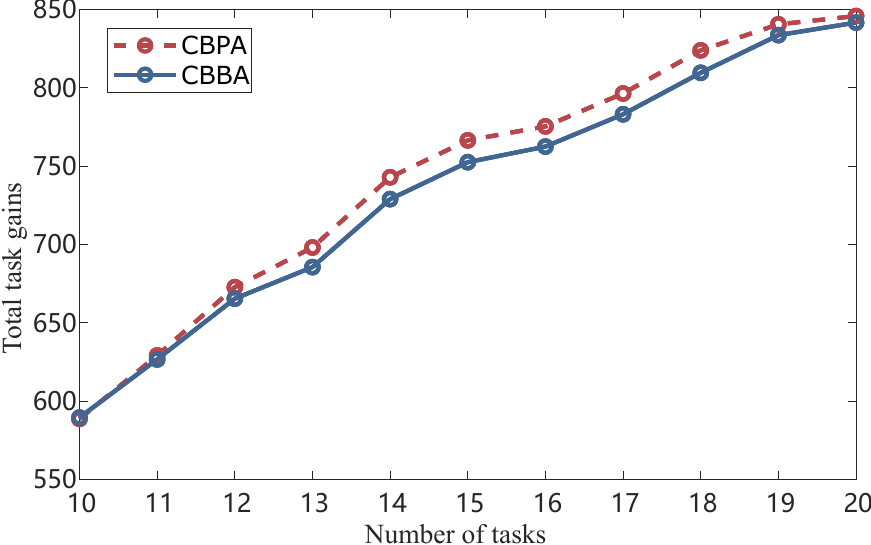}
\caption{Task Total Gains}
\label{F5}
\vspace{-0.5cm}
\end{figure}

Since the CBBA considers the single robot task, Case 2 is designed with five strike robots executing different numbers of strike tasks.
Each task has a static gain of 100 and a strike requirement value of 30. The time discount factor is 0.01, and each robot carries a strike payload value of 100.
There are 11 groups of experiments, with the number of tasks increased from 10 to 20. Each group performed an average of 20 independent experiments. The results of the total task gains are shown in Fig.\ref{F5}.

The results show that a single robot appears to have insufficient striking capacity when the number of tasks exceeds 10. CBPA has a higher total task gains and resource utilization than CBBA. This means that CBPA is able to make better use of robot payloads and has excellent robot resource utilization within robots' capability.

\section{Conclusion}  
In this paper, a consensus-based payload algorithm was presented to ensure that a robot coalition with matched capabilities executes each task for the multi-robot task allocation problem of robot payload consumption.
Physical and simulation experiments demonstrate that CBPA enhances robot resource utilization and effectively utilizes robot-carried payloads. In future work, the algorithm should be improved to enhance fault tolerance for better application in real multi-robot system.

\footnotesize
\bibliographystyle{IEEEtran}
\bibliography{IEEEref}

\end{document}